\documentclass[letterpaper]{article} 
\usepackage{aaai25}  
\usepackage{times}  
\usepackage{helvet}  
\usepackage{courier}  
\usepackage[hyphens]{url}  
\usepackage{graphicx} 
\usepackage{natbib}  
\usepackage{caption} 
\usepackage{algorithm}
\usepackage{algorithmic}

\usepackage{booktabs}
\usepackage{multirow}

\usepackage{amsthm}
\newtheorem{theorem}{Theorem}
\usepackage{amsmath}  
\usepackage{amssymb}  
\usepackage{amsfonts} 
\usepackage{xcolor}   
\usepackage{newfloat}
\usepackage{listings}
\DeclareCaptionStyle{ruled}{labelfont=normalfont,labelsep=colon,strut=off} 
\floatstyle{ruled}
\newfloat{listing}{tb}{lst}{}
\floatname{listing}{Listing}
\title{Enhancing Adversarial Attacks via Parameter Adaptive Adversarial Attack}
\author {
    Zhibo Jin\textsuperscript{\rm 1}\thanks{These authors contributed equally to this work.},
    Jiayu Zhang\textsuperscript{\rm 2}\footnotemark[1],
    Zhiyu Zhu\textsuperscript{\rm 1},
    Chenyu Zhang,\\
    Jiahao Huang\textsuperscript{\rm 1},
    Jianlong Zhou\textsuperscript{\rm 1},
    and Fang Chen\textsuperscript{\rm 1}
}

\affiliations {
    \textsuperscript{\rm 1}The University of Technology Sydney\\
    \textsuperscript{\rm 2}Suzhou Yierqi\\
}

\usepackage{bibentry}

\begin{document}

\maketitle

\begin{abstract}
In recent times, the swift evolution of adversarial attacks has captured widespread attention, particularly concerning their transferability and other performance attributes. These techniques are primarily executed at the sample level, frequently overlooking the intrinsic parameters of models. Such neglect suggests that the perturbations introduced in adversarial samples might have the potential for further reduction. Given the essence of adversarial attacks is to impair model integrity with minimal noise on original samples, exploring avenues to maximize the utility of such perturbations is imperative. Against this backdrop, we have delved into the complexities of adversarial attack algorithms, dissecting the adversarial process into two critical phases: the Directional Supervision Process (DSP) and the Directional Optimization Process (DOP). While DSP determines the direction of updates based on the current samples and model parameters, it has been observed that existing model parameters may not always be conducive to adversarial attacks. The impact of models on adversarial efficacy is often overlooked in current research, leading to the neglect of DSP. We propose that under certain conditions, fine-tuning model parameters can significantly enhance the quality of DSP. For the first time, we propose that under certain conditions, fine-tuning model parameters can significantly improve the quality of the DSP. We provide, for the first time, rigorous mathematical definitions and proofs for these conditions, and introduce multiple methods for fine-tuning model parameters within DSP. Our extensive experiments substantiate the effectiveness of the proposed P3A method. Our code is accessible at: \url{https://anonymous.4open.science/r/P3A-A12C/}
\end{abstract}

%

\section{Introduction}
\label{sec:intro}

Adversarial attacks hold a pivotal role in the domain of artificial intelligence security, particularly within the context of deep learning models. As artificial intelligence technologies find extensive applications ranging from autonomous driving~\cite{kong2020physgan} and facial recognition~\cite{massoli2021detection} to financial fraud detection~\cite{fursov2021adversarial} and health diagnostics~\cite{ghaffari2022adversarial}, the robustness of models becomes increasingly paramount. Adversarial attacks, by exposing the vulnerabilities of models when confronted with meticulously crafted inputs, underscore the imperative of fortifying models against such incursions. Presently, a significant portion of the research is devoted to evaluating the performance of adversarial attacks in a black-box setting, where the attacker is precluded from accessing or obtaining information about the internals of the target model. This focus is particularly pertinent, as black-box attack scenarios more closely align with the practical contexts encountered in real-world applications. However, it is noteworthy that the enthusiasm for research into white-box adversarial attacks, where attackers have complete knowledge of the model, has seen a gradual decline in recent years.

Two potential reasons might account for this phenomenon. Firstly, the high success rates achieved by white-box approaches could have led some researchers to overlook aspects beyond performance metrics. In other words, a high success rate in white-box attacks does not necessarily imply that the domain of white-box adversarial attacks has been exhaustively explored. Secondly, the unique circumstances, which need to fully access the information of target models, surrounding white-box attacks may lead some researchers to perceive them as less applicable in real-world scenarios, potentially diminishing their practicality and deterring further in-depth investigation.

However, it is precisely because attackers possess complete knowledge of the target model, including its architecture and parameters, in white-box adversarial attack scenarios, that they can fully leverage this information to make the attacks more precise and effective. Consequently, for researchers with access to model information, this context not only reveals potential vulnerabilities of the models but also provides a valuable opportunity to enhance model robustness~\cite{akhtar2021advances}. By thoroughly investigating and developing stronger adversarial attack techniques, such as exploring black-box attacks which we have also demonstrated in our experiments, we can not only gain a better understanding of the limitations of models but also advance research in robust defense mechanisms.

Current optimization strategies for adversarial attacks, such as MI-FGSM~\cite{dong2018boosting}, DI-FGSM~\cite{xie2019improving}, and SINI-FGSM~\cite{lin2019nesterov}, primarily focus on attacking the model at the sample level, often overlooking the influence of the model's inherent characteristics on the attack's effectiveness. By delving into the fundamental principles of adversarial attacks, we recognize that certain regions of a model may be more prone to attacks, while others may not, and that the model's landscape is highly non-uniform. Leveraging this insight, we propose a methodology to effectively modify the target model's parameters, making it more susceptible to attacks and, consequently, enhancing the efficacy of adversarial attack samples. This approach serves a dual purpose: firstly, to augment the effectiveness of adversarial attacks by, for the first time, introducing and substantiating the impact of the model's characteristics on such attacks; and secondly, to uncover the model's vulnerabilities. If a model becomes more vulnerable post-modification, it could suggest that updates leaning towards increased susceptibility might induce greater instability, thus facilitating an assessment of the model's inherent robustness traits.

\begin{figure}
  \centering
  \includegraphics[width=\linewidth]{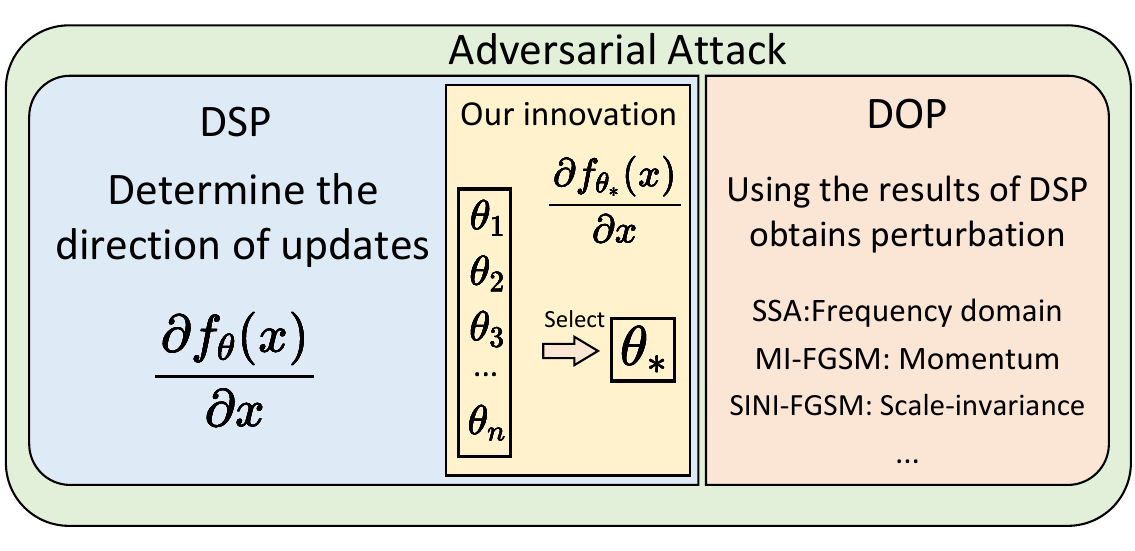}
  \caption{Diagram of Adversarial Attack Process with DSP and DOP Components}
  \label{flowchart}
\end{figure}

Given that the objective of adversarial attacks is to compromise the model's integrity with minimal perturbation to the original samples, it is imperative to study how to maximize the utilization of such disturbances. As shown in Fig.~\ref{flowchart}, we introduce two key steps in current adversarial attack algorithms: the Directional Supervision Process (DSP) and the Directional Optimization Process (DOP).

The DSP determines the update direction for adversarial samples, and we will discuss its impact on adversarial attacks in subsequent text. The DOP, following DSP, involves adjusting the update direction using optimization techniques, such as the frequency domain in Spectrum Simulation Attack (SSA)~\cite{long2022frequency}, the momentum mechanism in MI-FGSM, transformation operations in DI-FGSM, and scaling in SINI-FGSM. We detail the formulas for both steps in \textit{Section DSP and DOP in Adversarial Attacks}

Typically, the DSP step determines the update direction based on the current parameters of the model. However, we identify that the existing model parameters are not necessarily optimal for adversarial attacks. Minor adjustments to the model parameters can significantly enhance the quality of DSP under certain conditions. We derive the impact of model parameter fine-tuning on the DSP step and propose four methods of parameter adjustment tailored to the current adversarial samples. The selection of the adjustment method is based on the estimated impact. We refer to this algorithm as Parameter Adaptive adversarial attack (P3A). Our contributions are as follows:
\begin{enumerate}
    \item By delving into the fundamental principles of adversarial attacks, we categorize the adversarial process into two distinct phases: the Directional Supervision Process (DSP) and the Directional Optimization Process (DOP).
    \item We define adversarial attacks and derive the influence of parameter fine-tuning on these attacks, providing a detailed proof process.
    \item We propose four parameter fine-tuning techniques that can affect the outcome of adversarial attacks. By combining these with an estimate of their impact, we facilitate the selection of the most appropriate adjustment method.
    \item We commit to open sourcing our approach and conducting extensive experiments to validate its effectiveness.
\end{enumerate}

\section{Related work}

\subsection{White-box Attacks}
In the field of adversarial machine learning, white-box adversarial attacks have progressively exposed the vulnerabilities of deep neural networks to carefully crafted inputs through continuous iteration and innovation. From the initial Fast Gradient Sign Method (FGSM)~\cite{goodfellow2014explaining} to the more refined Projected Gradient Descent (PGD)~\cite{madry2017towards} and Carlini \& Wagner Attack (C\&W Attack)~\cite{carlini2017towards}, each method has deepened our understanding of adversarial samples and model robustness based on the insights gained from its predecessors.

Initially, the FGSM pioneered the generation of adversarial samples using model gradient information, laying the theoretical groundwork for the DSP mechanism, which we discuss in more detail in \textit{Section DSP and DOP in Adversarial Attacks}. Goodfellow et al.~\cite{goodfellow2014explaining} demonstrated how the linear properties of deep networks could be exploited for effective attacks through a single substantial gradient update, generating adversarial samples in a simple yet direct manner. The Basic Iterative Method (BIM)~\cite{kurakin2018adversarial} improved upon FGSM by introducing an iterative process for gradually adjusting adversarial samples. Kurakin et al.~\cite{kurakin2018adversarial} refined the generation of adversarial samples to more precisely target the model's vulnerabilities through this meticulous iterative process, thereby enhancing the effectiveness of the attacks. The PGD~\cite{madry2017towards}, a further extension of BIM, involved projecting perturbations at each step to ensure that the adversarial samples remained within a predefined disturbance range. This constraint made the adversarial samples generated by PGD both potent and practical, especially effective for evaluating model robustness. Additionally, the C\&W Attack further increased the stealthiness of adversarial samples. Carlini and Wagner~\cite{carlini2017towards} generated almost imperceptible perturbations through a carefully designed optimization problem, showcasing the potential of adversarial attacks in terms of stealthiness and simultaneously raising the bar for defense mechanisms.

\subsection{Black-box Attacks}
In the realm of black-box attacks, various methods have been devised to enhance the transferability and effectiveness of adversarial examples across different models. The Translation-Invariant Attack Method (TI-FGSM)~\cite{dong2019evading} optimizes adversarial examples using a set of translated images to reduce sensitivity to the discriminative regions of the target model, thereby improving transferability. Similarly, the Diverse Input Iterative Fast Gradient Sign Method (DI-FGSM)~\cite{xie2019improving} introduces input diversity through random, differentiable transformations at each iteration, aiming to produce adversarial examples effective against a broader range of models. Likewise, the Large Geometric Vicinity (LGV)~\cite{gubri2022lgv} approach also aggregates multiple gradient information to produce more effective adversarial samples by simulating model parameters from some training datasets, and then targeting these parameters for attacks to generate more transferable adversarial samples. The Momentum Iterative Fast Gradient Sign Method (MI-FGSM)~\cite{dong2018boosting} incorporates a momentum term to stabilize update directions and escape suboptimal maxima, enhancing the efficacy of attacks, especially in black-box scenarios. The SINI-FGSM~\cite{lin2019nesterov} method merges Nesterov accelerated gradients and scale-invariance to further refine adversarial example transferability by optimizing perturbations over scale variations and looking ahead in the optimization process. Additionally, the SSA~\cite{long2022frequency} employs frequency domain transformations to simulate a variety of models, effectively bridging the gap between the substitute model used for attack generation and the victim model, thus elevating the success rate of adversarial attacks across diverse models. In summary, These methodologies primarily focus on optimization and innovation during the DOP process, often neglecting the role of DSP, which we discuss in more detail in \textit{Section DSP and DOP in Adversarial Attacks}.

\subsection{Adversarial Defenses}
Adversarial training and ensemble defenses are advanced strategies to improve machine learning model resilience against adversarial attacks~\cite{lu2022ensemble}. Adversarial training involves integrating adversarial examples into the training set, enabling models to learn from these perturbations and build resistance. This method transforms potential vulnerabilities into strengths by familiarizing the model with various attack vectors. Ensemble defenses, conversely, utilize diversity by combining different models or methods for collective decision-making~\cite{tramer2017ensemble}. This approach introduces redundancy, mitigating the impact of individual model failures and decreasing the overall success rate of attacks. The ensemble strategy benefits from the distinct sensitivities and strengths of its components, complicating the task for adversaries to devise attacks that simultaneously deceive all models~\cite{deng2024understanding}. Adversarial training and ensemble defenses together form a robust defense system that leverages adversarial insights and the collective robustness of diverse mechanisms, offering comprehensive protection against the dynamic nature of adversarial threats~\cite{croce2020robustbench}.

\section{Method}
\subsection{Problem Definition}

Consider a deep learning model $f$ parameterized by $\theta$, which maps an input sample $x$ to an output prediction $y = f(x; \theta)$, where $x \in \mathbb{R}^{m}$ and $\theta \in \mathbb{R}^n$. In the context of adversarial attacks, the objective is to craft a perturbation $\delta$ such that the perturbed input $x_{adv} = x + \delta$ causes the model to output an incorrect prediction $f(x_{adv}; \theta) \neq y_{true}$, where $y_{true}$ is the true label of $x$.

The loss function is utilized to measure the distance between the model's output distribution and the true distribution, hence, a larger loss function value indicates a more effective adversarial attack. The adversarial perturbation $\delta$ is generated to maximize the loss function $L(f(x_{adv}; \theta), y_{true})$, which in Sec.~\ref{P3A_proof} is denoted as $L_{fy} (x,\theta)$, while conforming to a constraint on the perturbation's magnitude to ensure imperceptibility. This is typically quantified by some norm $\|\delta\|_p \leq \epsilon$, where $\epsilon$ is a small constant.

Formally, the problem can be stated as the following optimization problem:
\begin{equation}
    \max_{\delta} L(f(x + \delta; \theta), y_{true}), \quad \text{s.t.} \quad \|\delta\|_p \leq \epsilon.
\end{equation}

In this work, we extend the conventional adversarial attack framework to incorporate the influence of model parameters $\theta$ in crafting adversarial samples. Specifically, we examine the role of the first-order derivatives $\frac{\partial L}{\partial x}$ and $\frac{\partial L}{\partial \theta}$, which represent the sensitivities of the loss function to changes in the input sample and model parameters, respectively. Furthermore, we consider the second-order partial derivative $\frac{\partial^2 L}{\partial x \partial \theta}$, which assesses the impact of variations in model parameters $\theta$ on the gradient of the loss function with respect to the input $x$, thereby offering a refined direction for crafting more effective adversarial samples.

\subsection{DSP and DOP in Adversarial Attacks}\label{DSP_detail}

In the following section on DSP and DOP in adversarial attacks, we will delve into the mechanisms and mathematical formulations that underpin the generation of adversarial examples. The \textbf{DSP} refers to the methodology by which the direction of the perturbation is determined, often leveraging the gradient of the loss function with respect to the input data. The \textbf{DOP}, on the other hand, involves the refinement and application of this direction to iteratively update the adversarial example, ensuring that the perturbation is optimized to achieve the desired effect of misleading the model while remaining imperceptible or minimal.

We will explore several key adversarial attack strategies, including the Fast Gradient Sign Method (FGSM), Projected Gradient Descent (PGD), Momentum Iterative Fast Gradient Sign Method (MI-FGSM), Diverse Input Fast Gradient Sign Method (DI-FGSM), and Scale-Invariant Iterative Fast Gradient Sign Method (SINI-FGSM). Each of these methods employs variations of DSP and DOP to manipulate the input data in a way that maximizes the loss of the targeted model, thereby compromising its performance. Through mathematical formulations, we will illustrate how these attacks generate adversarial examples, highlighting the role of gradient calculation in determining the direction of perturbation and the iterative optimization process that fine-tunes these perturbations. We have analyzed in detail in the appendix how the DSP and DOP components constitute adversarial attack methods.

We divide the process of adversarial attacks into two parts: DSP and DOP. The purpose of DSP is to provide directional supervision for the gradient ascent process of adversarial samples. In this context, FGSM serves as a clear example, demonstrating how to determine the direction of adversarial perturbation by leveraging the gradient of the loss function with respect to the input, specifically by computing the gradient of the input with respect to the loss function and applying the sign function to dictate the optimal direction of perturbation. In subsequent sections, we will discuss the impact of model parameters on DSP and derive the influence of changes in model parameters on the outcomes of gradient ascent. Through theoretical derivation, we can achieve more optimal DSP results by fine-tuning the current parameters.

\subsection{Second-Order Partial Derivatives and DFD}
The objective of adversarial attacks is to maximize the loss function by adding limited perturbations to the samples. It is crucial to examine the relationship between changes in each dimension of the samples and the loss function value (the partial derivative of the loss function with respect to the input samples, $\frac{\partial L}{\partial x}$) to identify the appropriate direction for sample updates. If we consider the impact of model parameters, we need to compute the second-order partial derivatives $\frac{\partial^2 L}{\partial x \partial \theta}$. According to Clairaut's theorem~\cite{james1966advanced}, $\frac{\partial^2 L}{\partial x \partial \theta} = \frac{\partial^2 L^\top}{\partial x \partial \theta}$, meaning the order of input samples and model parameters can be interchanged without affecting the proof process. Due to the high computational cost and storage requirement of calculating second-order partial derivatives $\frac{\partial^2 L}{\partial x \partial \theta} \in \mathbb{R}^{m\times n}$, an approximation of the calculation process can be made by estimating the influence of $\theta$ on $\frac{\partial L}{\partial x}$. Here, we present the theorem in its general form:

\begin{theorem}[Directional Finite Difference (DFD)]
The second-order derivative of a function can be estimated using the difference in first-order derivatives. Given a function \(f(v,u)\) with two variables, we have
\begin{equation}
\begin{aligned}
\small
\frac{\partial^2 f(v, u)}{\partial v \partial u} \Delta u & \approx \frac{\left(\frac{\partial f(v, u+\varepsilon \Delta u)}{\partial v}-\frac{\partial f(v, u)}{\partial v}\right)}{\varepsilon} \\
&\approx \frac{\frac{\partial f(v, u+\varepsilon \Delta u)}{\partial v}-\frac{\partial f(v, u-\varepsilon \Delta u)}{\partial v}}{2 \varepsilon}
\end{aligned}
\end{equation}
\end{theorem}

\noindent\textit{Proof 1}.

\begin{equation}
\small
\frac{\partial f(v, u+\varepsilon \Delta u)}{\partial v}=\frac{\partial f(v, u)}{\partial v}+\varepsilon \cdot \frac{\partial^2 f(v,u)}{\partial v\partial  u} \cdot \Delta u+ \mathcal{O}(\varepsilon^2) 
\end{equation}
The proof follows by rearranging the terms.

In this context, the input sample $x$ can be regarded as $v$, and the parameter $\theta$ is $u$. Directional finite difference serves as an effective tool for assessing the higher-order (second-order in this paper) properties of a function.

\subsection{Parameter Adaptive Adversarial Attack} \label{P3A_proof}

Utilizing the definition and proof process of directional finite difference, it becomes evident that during estimation, the change in variable $u$, denoted as $\Delta u$, needs to be specified. In neural networks, $\Delta u$ corresponds to changes in parameters $\Delta \theta$. We then define the impact of parameter changes on adversarial effectiveness:
\begin{theorem}Given model parameters $\theta$, we have

$ L_{\theta^{\prime}} \geqslant L_{\theta} \quad \text{s.t.} \quad (g_{\theta}^{\prime} - g_{\theta})^\top \cdot g_{\theta} \geqslant 0, \theta^{\prime} = \theta + \alpha \cdot F(\theta)$


\begin{equation}
\small
    L_{\theta} = L_{fy}\left(x + \beta \frac{\partial L_{fy}(x, \theta)}{\partial x}, \theta\right)
\end{equation}
\begin{equation}
\small
    L_{\theta^{\prime}} = L_{fy}\left(x + \beta \frac{\partial L_{fy}(x, \theta + \alpha \cdot F(\theta))}{\partial x}, \theta\right)
\end{equation}

\end{theorem}

where $\cdot$ denoted as dot production, $g_{\theta} = \frac{\partial L_{fy}(x, \theta)}{\partial x}$ represents the gradient information of the neural network's loss function calculated for the current sample under the effect of parameters $\theta$, and $\Delta \theta = \alpha F(\theta)$ represents the value of parameter change.


{\small
\textit{Proof 2}.
\begin{equation}
\small
    L_{\theta}^{\prime} = L_{fy}\left(x + \beta \frac{\partial L_{fy}(x, \theta + \alpha \cdot F(\theta))}{\partial x}, \theta\right)
\end{equation}
\begin{equation}
\small
    = L_{fy}(x, \theta) + \beta \frac{\partial L_{fy}(x, \theta + \alpha \cdot F(\theta))^\top}{\partial x} \cdot \frac{\partial L_{fy}(x, \theta)}{\partial x} + \mathcal{O}_1
\end{equation}

\begin{equation}\label{eqn:1}
\begin{aligned}
\small
    & = L_{fy}(x, \theta) + \beta \left[\frac{\partial L_{fy}(x, \theta)}{\partial x} + \alpha \cdot F(\theta)^\top \cdot \frac{\partial^2 L_{fy}(x, \theta)}{\partial x \partial \theta} + \mathcal{O}_2\right]^\top \\
     & \cdot \frac{\partial L_{fy}(x, \theta)}{\partial x} + \mathcal{O}_1 
\end{aligned}
\end{equation}

\begin{equation}
\begin{aligned}
\label{equ:16}
\small
    & \approx \underbrace{L_{fy}(x, \theta) + \beta \cdot \frac{\partial L_{fy}(x, \theta)^\top}{\partial x} \cdot \frac{\partial L_{fy}(x, \theta)}{\partial x}}_{L_{\theta}} \\
    & + \beta \cdot \alpha \cdot \frac{\partial^2 L_{fy}(x, \theta)^\top}{\partial x \partial \theta} \cdot F(\theta) \cdot \frac{\partial L_{fy}(x, \theta)}{\partial x}
\end{aligned}
\end{equation}
}
From Theorem 1, we have
{\small
\begin{equation}
\begin{aligned}
&\alpha \frac{\partial^2 L_{fy}(x, \theta)^\top}{\partial x \partial \theta} \cdot F(\theta) \\ &\approx \frac{1}{\alpha} \cdot \left(\frac{\partial L_{fy}(x, \theta + \alpha \cdot F(\theta))}{\partial x} - \frac{\partial L_{fy}(x, \theta)}{\partial x}\right) \\
&\propto \underbrace{\frac{\partial L_{fy}(x, \theta + \alpha \cdot F(\theta))}{\partial x}}_{g_{\theta}^{\prime}} - \underbrace{\frac{\partial L_{fy}(x, \theta)}{\partial x}}_{g_{\theta}} \label{equ:18}
\end{aligned}
\end{equation}
}
\noindent By combining the above equation~\ref{equ:16} and equation~\ref{equ:18}, Theorem 2 is proven.

Where $\mathcal{O}_1$ and $\mathcal{O}_2$ represent higher-order infinitesimals. The gradient of $x$ with respect to $\theta$ after fine-tuning, denoted as $g_{\theta}^{\prime}$, and the gradient of $x$ with respect to $\theta$ before fine-tuning, denoted as $g_{\theta}$.

\begin{theorem}Given model parameters $\theta$, we have
{\small$L_{\theta^{\prime}} \geqslant L_{\theta} \quad \text{s.t.} \quad (g_{\theta}^{\prime} - g_{\theta}) \cdot g_{\theta} \leqslant 0, \quad \theta^{\prime} = \theta - \alpha \cdot F(\theta)$}
\end{theorem}

We then present four potential methods for parameter updates:
\begin{enumerate} \label{updates_methods}
    \item \textbf{Defensive Perspective}: $\theta = \theta \pm  \alpha \frac{\partial L_{y}(x, \theta)}{\partial \theta}$
    \item \textbf{Uniform Method}: (most general, treating each parameter equitably): $\theta = \theta \pm  \alpha \cdot S$, where $S$ is a constant value.
    \item \textbf{Magnitude Consideration} (an extension of the uniform approach): $\theta = \theta \pm  \alpha \cdot \frac{\partial \theta^2}{\partial \theta}$, taking into account the magnitude of parameters. In this scheme, parameters with larger values are explored more extensively, as they are typically more significant in neural networks.
    \item \textbf{Decoupling Method}: $\theta = \theta \pm  \alpha \cdot \operatorname{sign}\left(\frac{\partial L_{fy}(x, \theta)}{\partial \theta}\right)$, employs the sign function to facilitate the decoupling of adversarial attacks~\cite{goodfellow2014explaining}, utilizing an adversarial strategy for parameter exploration. Parameters with larger gradients are subject to more extensive exploration, as a larger gradient often implies a greater impact on the outcome, signifying the parameter's importance.
\end{enumerate}

Method 2 represents the most fundamental update strategy, where each parameter is incremented by a fixed value, ensuring fairness. Method 3 builds on this by considering the size of the parameters, allowing for more exploration of parameters with larger values, which are often more critical in neural networks. Methods 1 and 4 adopt an adversarial approach to parameter exploration, focusing more on parameters with larger gradients. Additionally, performing gradient ascent or descent on parameters can be viewed as adapting model parameters to current attack targets within an adversarial defense context. Throughout the experimental process, we employed these four methods to update model parameters $\theta$, and utilized Theorems 2 and 3 to evaluate whether the modified parameters were conducive to the DSP process. Subsequently, the selected model parameters were substituted for the original parameters for the DSP process. 

\section{Experiments}
\label{sec:exper}

\subsection{Experimental Setup}

\subsubsection{Dataset}
Following the protocol established in previous studies~\cite{zhang2022improving,long2022frequency}, we utilized the same dataset for our experimentation. Specifically, we randomly selected 1000 images from the ILSVRC 2012 validation set~\cite{ILSVRC15} for testing purposes, ensuring consistency in experimental conditions across different studies.

\subsubsection{Models}
Our study rigorously assesses various CNN architectures and cutting-edge vision Transformer models to demonstrate the widespread effectiveness of our P3A method. We examined Inception models (Inc-v3~\cite{szegedy2016rethinking}, Inc-v4~\cite{szegedy2017inception}, Inc-Res-v2~\cite{szegedy2017inception}), ResNet~\cite{he2016deep} variants (Res-50, Res-101, Res-152), and adversarially trained models (Inc-v3-adv~\cite{kurakin2016adversarial}, Inc-v3-adv-ens3, Inc-v3-adv-ens4, Inc-Res-ens~\cite{tramer2017ensemble}) to evaluate defense mechanisms. Additionally, we included Transformer-based models ViT-B/16~\cite{dosovitskiy2020image} and MaxViT-T~\cite{tu2022maxvit} to assess the robustness of contemporary vision models against adversarial threats.

\subsubsection{Baseline Methods}
Our experimental design for the white-box attack scenario amalgamates six widely recognized attack techniques: BIM~\cite{kurakin2018adversarial}, PGD~\cite{madry2017towards}, DI-FGSM~\cite{xie2019improving}, TI-FGSM~\cite{dong2019evading}, MI-FGSM~\cite{dong2018boosting}, and SINI-FGSM~\cite{lin2019nesterov}. This comprehensive approach is employed to underscore the versatility of our P3A method in enhancing white-box attack efficacy across various methodologies. Conversely, for the transfer attack scenario, our focus shifts to assessing models' resilience against transferable attacks. Accordingly, we selected two transfer attack methods published at top-tier academic conferences since 2023: Momentum Integrated Gradients (MIG)~\cite{ma2023transferable}, and Gradient Relevance Attack (GRA)~\cite{zhu2023boosting}, to validate the efficacy of our approach in these contexts as well.

\subsubsection{Metrics}
Our evaluation utilized two principal metrics. Firstly, the change in model Loss, which is crucial for assessing white-box adversarial attacks aiming to maximize this Loss. A substantial Loss increase indicates the attack's efficacy in diverting model predictions from true labels, showcasing the impact level on the model. Secondly, the Attack Success Rate (ASR), defined as the ratio of test samples that led to incorrect model predictions, was used. ASR directly assesses an attack's ability to undermine model accuracy, serving as a straightforward measure of attack effectiveness in both white-box and black-box scenarios.

\subsubsection{Parameters}
In the white-box attack scenario, the hyperparameters for our P3A method include the attack step size $\alpha$ and the learning rate for fine-tuning model parameters. We set the attack step size $\alpha$ to 1 and the learning rate for fine-tuning model parameters to 0.0001. For the comparative experiments, all parameter settings adhered to the standard configurations established by existing transfer attack benchmarks~\cite{jin2024short}.

\subsection{Results}
\subsubsection{White-box Scenario}
\begin{figure*}[t]
    \centering
    \includegraphics[width=0.7\linewidth]{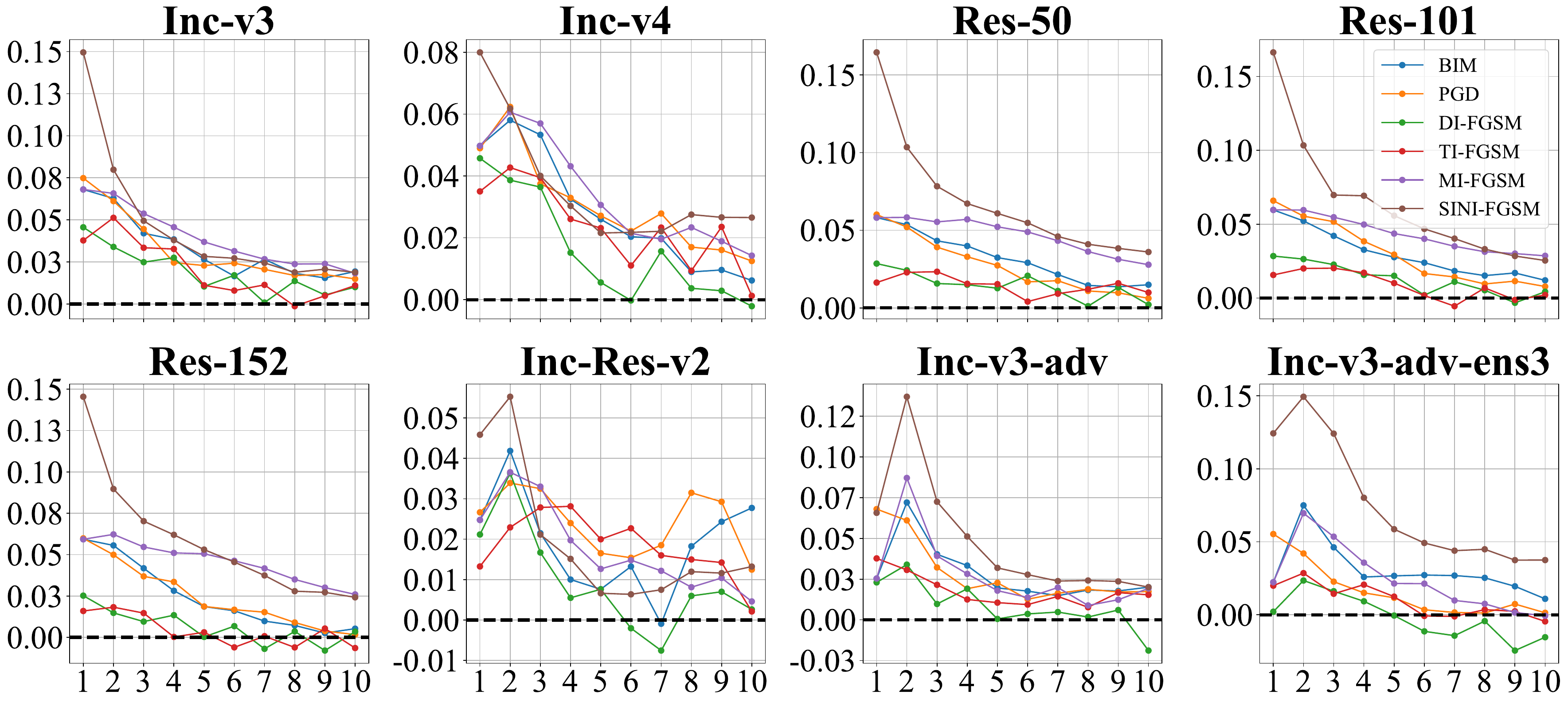}
    \caption{Incremental Efficacy of the P3A Method on Cross Entropy Loss Across Attack Steps. The y-axis represents the change in cross-entropy loss, and the x-axis represents attack steps. The black dashed line means the efficacy of attack methods without P3A.}
    \label{fig:CE_loss}
\end{figure*}

\begin{figure*}[t]
    \centering
    \includegraphics[width=0.7\linewidth]{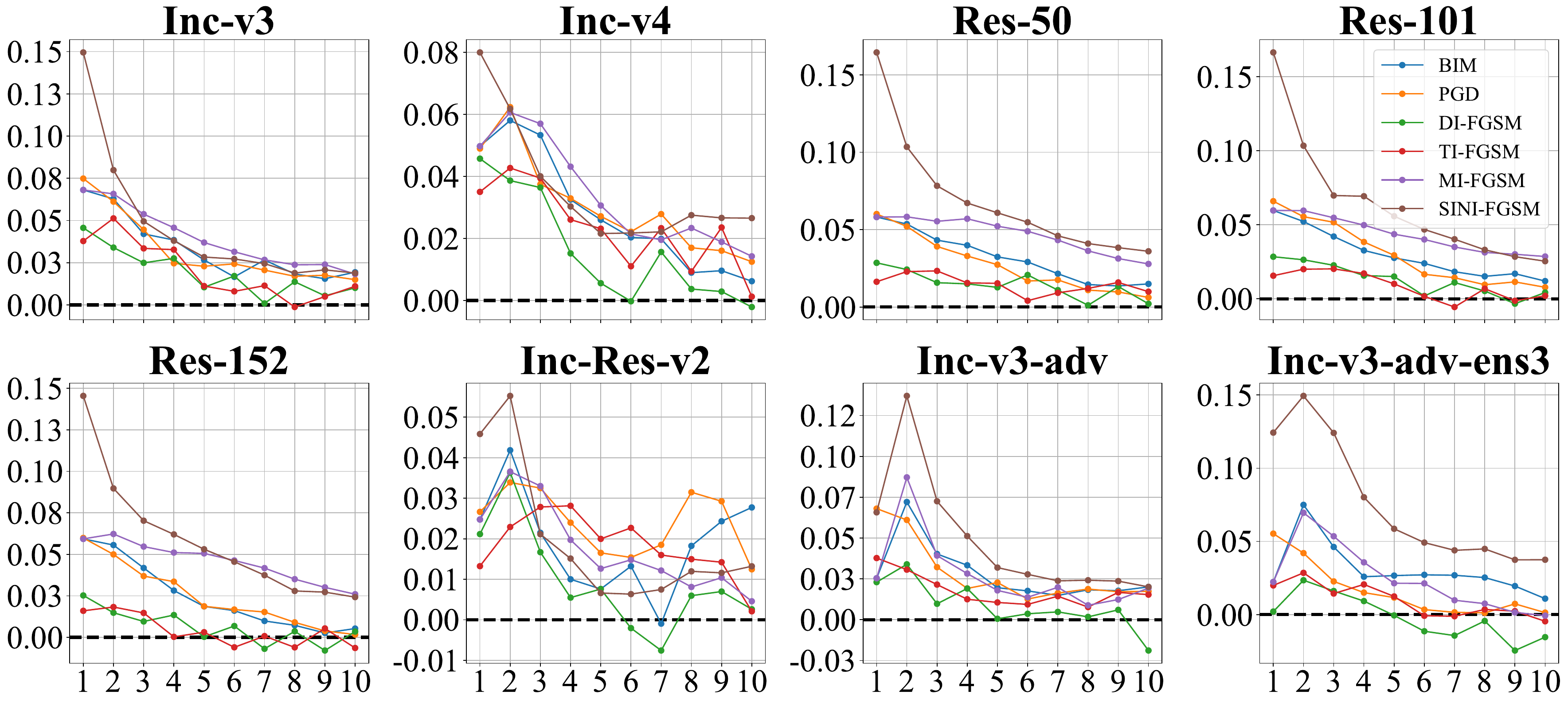}
    \caption{Incremental Efficacy of the P3A Method on KL Loss Across Attack Steps. The y-axis represents the change in KL loss, and the x-axis represents attack steps. The black dashed line means the efficacy of attack methods without P3A.}
    \label{fig:KL_Loss}
\end{figure*}

\begin{figure*}[t]
    \centering
    \includegraphics[width=0.7\linewidth]{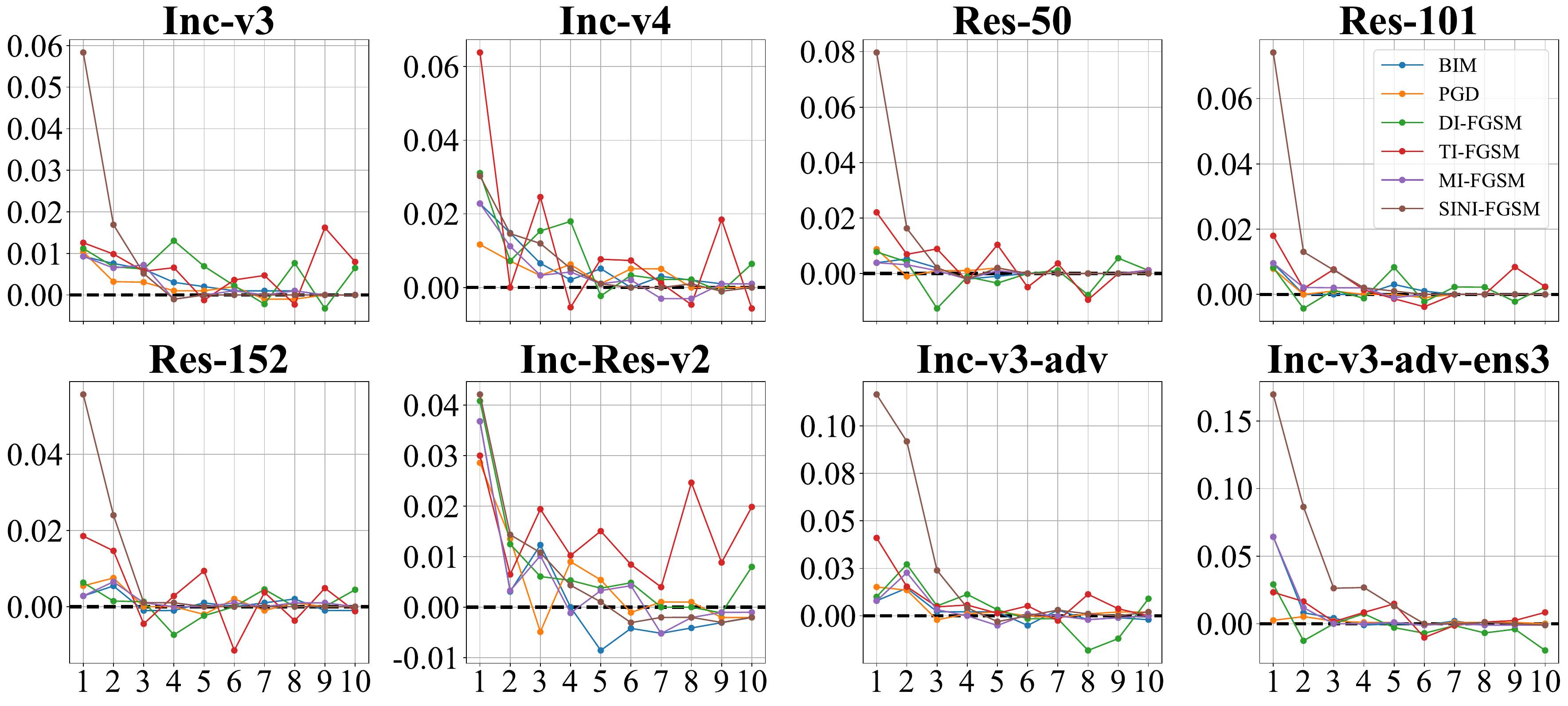}
    \caption{Incremental Efficacy of the P3A Method on ASR Across Attack Steps. The y-axis represents the change in ASR, and the x-axis represents attack steps. The black dashed line means the efficacy of attack methods without P3A.}
    \label{fig:asr}
\end{figure*}

\begin{table*}[t]
\centering
\caption{Comparison of ASR, showcasing the performance of P3A against other state-of-the-art transferable attack methods across different models. The best performance results are marked in bold.}
\label{table:transfer_attack_comparison}
\resizebox{.9\linewidth}{!}{%
\begin{tabular}{@{}c|c|ccccccccccc|c@{}}
\toprule
Model                     & Method & Inc-v3          & Inc-v4          & IncRes-v2       & Res-50          & Res-101         & Res-152         & Inc-v3-ens3     & Inc-v3-ens4     & IncRes-ens      & ViT-B/16        & MaxViT-T        & Average         \\ \midrule
\multirow{3}{*}{Inc-v3}   & MIG    & -               & 70.6\%          & 69.1\%          & 69.2\%          & 64.5\%          & 63.5\%          & 38.9\%          & 40.2\%          & 22.0\%          & 32.1\%          & 31.5\%          & 50.2\%          \\
                          & GRA    & -               & 82.0\%          & 81.4\%          & 76.8\%          & 74.0\%          & 75.1\%          & 53.1\%          & 50.6\%          & 29.9\%          & 39.4\%          & 43.0\%          & 60.5\%          \\
                          & P3A(our)    & -               & \textbf{89.0\%} & \textbf{87.4\%} & \textbf{81.6\%} & \textbf{81.2\%} & \textbf{81.1\%} & \textbf{74.2\%} & \textbf{73.4\%} & \textbf{59.3\%} & \textbf{54.3\%} & \textbf{47.9\%} & \textbf{72.9\%} \\ \midrule
\multirow{3}{*}{Inc-v4}   & MIG    & 86.3\%          & -               & 78.6\%          & 77.7\%          & 73.3\%          & 74.8\%          & 52.7\%          & 48.2\%          & 33.8\%          & 40.2\%          & 46.9\%          & 61.3\%          \\
                          & GRA    & 86.8\%          & -               & 80.8\%          & 77.7\%          & 75.1\%          & 76.1\%          & 54.6\%          & 51.6\%          & 34.2\%          & 40.5\%          & 49.7\%          & 62.7\%          \\
                          & P3A(our)    & \textbf{91.4\%} & -               & \textbf{87.7\%} & \textbf{83.1\%} & \textbf{81.2\%} & \textbf{82.3\%} & \textbf{75.4\%} & \textbf{72.9\%} & \textbf{63.6\%} & \textbf{57.7\%} & \textbf{56.0\%} & \textbf{75.1\%} \\ \midrule
\multirow{3}{*}{ViT-B/16} & MIG    & 67.8\%          & 62.4\%          & 55.6\%          & 65.5\%          & 62.6\%          & 61.4\%          & 57.6\%          & 56.8\%          & 50.3\%          & -               & 54.1\%          & 59.4\%          \\
                          & GRA    & 77.6\%          & 71.3\%          & 70.2\%          & 76.3\%          & 74.0\%          & 72.6\%          & 68.1\%          & 67.9\%          & 60.8\%          & -               & 72.4\%          & 71.1\%          \\
                          & P3A(our)    & \textbf{78.7\%} & \textbf{73.4\%} & \textbf{71.2\%} & \textbf{77.4\%} & \textbf{74.9\%} & \textbf{73.8\%} & \textbf{72.1\%} & \textbf{73.3\%} & \textbf{67.0\%} & -               & \textbf{63.3\%} & \textbf{72.5\%} \\ \midrule
\multirow{3}{*}{MaxViT-T} & MIG    & 74.2\%          & 75.8\%          & 67.7\%          & 73.4\%          & 69.4\%          & 69.1\%          & 55.7\%          & 57.0\%          & 50.1\%          & 63.1\%          & -               & 65.6\%          \\
                          & GRA    & 78.3\%          & 77.5\%          & 72.9\%          & 76.4\%          & 73.7\%          & 74.1\%          & 66.5\%          & 68.5\%          & 61.0\%          & 75.9\%          & -               & 72.5\%          \\
                          & P3A(our)    & \textbf{80.6\%} & \textbf{82.0\%} & \textbf{76.9\%} & \textbf{78.9\%} & \textbf{78.0\%} & \textbf{76.2\%} & \textbf{74.1\%} & \textbf{73.9\%} & \textbf{70.0\%} & \textbf{82.1\%} & -               & \textbf{77.3\%} \\ \bottomrule
\end{tabular}%
}
\end{table*}

\begin{figure*}[h!]
    \centering
    \includegraphics[width=\linewidth]{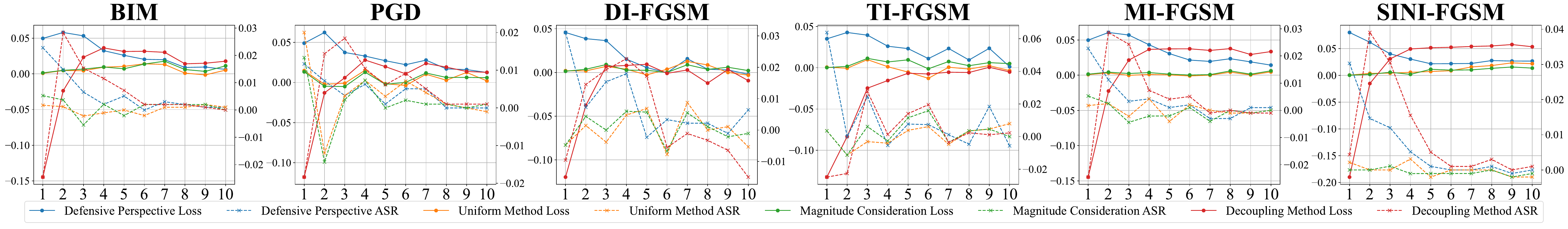}
    \caption{Performance of P3A on various models under 4 different parameter updates methods. The left y-axis represents the change in Cross Entropy Loss, and the right y-axis represents the change in ASR}
    \label{fig:compl}
\end{figure*}

In Fig.~\ref{fig:CE_loss} and Fig.~\ref{fig:KL_Loss}, we document the incremental efficacy of our P3A methodology with respect to the cross-entropy loss and KL Loss across a suite of attack methods, such as BIM, PGD, DI-FGSM, TI-FGSM, MI-FGSM, and SINI-FGSM. Notably, the P3A integration augments both Loss metrics across all attack strategies, signaling a consistent enhancement in attack effectiveness. Specifically, in comparison to the baseline methods, the cross-entropy Loss saw an average increase of 0.028, while the KL Loss averaged an increase of 0.031. The SINI-FGSM method demonstrated the most considerable escalation, with an average boost of 0.052 in cross-entropy Loss and 0.055 in KL Loss; the DI-FGSM method presented the least increase, enhancing both the cross-entropy Loss and KL Loss by 0.011. This signifies a pronounced compatibility and effectiveness of our approach with these particular types of attacks.

As the attack iterations progress, we witness a diminishing marginal increase in the Loss value. This trend can be attributed to the decreasing sensitivity of the model to further perturbations, reflective of the adaptive nature of deep learning models under sustained adversarial pressure. Nevertheless, the sustained incremental gains validate the potency and utility of the P3A method in enhancing adversarial attacks step-by-step.

Further examination reveals that adversarially trained models, such as Inc-v3-adv and Inc-v3-adv-ens3, also benefit from the P3A method, albeit to a lesser extent compared to models without adversarial hardening like Inc-v3, Inc-v4, and Res-50. This outcome suggests the efficacy of adversarial training in bolstering model robustness, and concurrently, the P3A method's capacity to introduce improvements even within robustified models.

Fig.~\ref{fig:asr} complements the findings of Fig.~\ref{fig:KL_Loss} by demonstrating the incremental impact of the P3A method on the ASR across attack steps for different models. The substantial increases in ASR, particularly evident in adversarially trained models Inc-v3-adv and Inc-v3-adv-ens3, underscore the adaptability and enhanced performance of the P3A method in bolstering attack success rates, especially within models fortified against such adversarial exploits.

\subsubsection{Black-box Scenario}

Table~\ref{table:transfer_attack_comparison} presents a comparison of the ASR between P3A and other state-of-the-art transferable attack methods across various models. In all tested models, P3A achieved the highest ASR in most cases, demonstrating its superiority in transferable attacks. Notably, P3A's performance was particularly outstanding when targeting models such as Inception-v3, Inception-v4, and MaxViT-T, with average ASRs reaching 72.9\%, 75.1\%, and 77.3\% respectively. This could be attributed to P3A's attack strategy effectively adapting to and exploiting the specific vulnerabilities of these models. Compared to the MIG and GRA methods, P3A exhibited significant improvements against adversarially robust models (e.g., Inc-v3-ens3 and Inc-v3-ens4). Even for more complex Transformer architectures (ViT-B/16 and MaxViT-T), P3A also demonstrated strong attack capabilities.

\subsection{Ablation Results}

\subsubsection{The Impact of 4 different parameter updates methods on P3A Performance}

In this section, we investigate the impact of the four different parameter update methods introduced in Sec.~\ref{updates_methods} on the performance of the P3A attack. The attack step size $\alpha$ was fixed at 1, and the learning rate (LR) was set to 0.0001. As shown in Fig.~\ref{fig:compl}, the experimental results indicate that among all the update methods compared, the performance of P3A is optimal when parameters are updated using the Defensive Perspective approach. 

\section{Conclusion}
In conclusion, this work presents a comprehensive examination of adversarial attacks through the lens of the Directional Supervision Process (DSP) and the Directional Optimization Process (DOP), delineating a nuanced approach to enhancing the efficacy of such attacks via Parameter Adaptive adversarial attack (P3A) methodology. By identifying the pivotal role of parameter fine-tuning and introducing four novel fine-tuning techniques, we demonstrate the potential for significant improvements in adversarial attack strategies. Our extensive experimental validation underscores the robustness and versatility of our approach across various models and attack vectors. Looking forward, we aim to explore the implications of extending the parameter exploration to more iterations and incorporating higher-order properties in the DSP, potentially unveiling deeper insights into model vulnerabilities and advancing the state of adversarial attack methodologies.

\bibliography{aaai25}

\clearpage
\appendix

\section{Analysis of DSP and DOP components}

In this section, we denote the \textbf{DSP} components of each method in {\color{red} red}, while the remaining parts pertain to \textbf{DOP}.

\textbf{FGSM}: This method generates adversarial examples by leveraging the sign of the gradient of the loss function with respect to the input, establishing the theoretical foundation for DSP.
\begin{equation}
\small
    x_{\text{adv}} = x + \epsilon \cdot \text{sign}\left({\color{red} \frac{\partial L(f(x, \theta), y_{\text{true}})}{\partial x}} \right)
\end{equation}

\textbf{PGD}: An iterative method that applies the FGSM multiple times with small step size and projects the adversarial examples into an $\epsilon$-ball around the original input.
\begin{equation}
\small
    x_{\text{adv}}^{t+1} = \Pi_{x+\epsilon}\left(x_{\text{adv}}^t + \alpha \cdot \text{sign}\left({\color{red} \frac{\partial L(f(x_{\text{adv}}^t, \theta), y_{\text{true}})}{\partial x}} \right)\right)
\end{equation}

\textbf{MI-FGSM}: Incorporates momentum into the iterative FGSM to stabilize update directions and escape from poor local maxima.
\begin{equation}
\small
    g^{t+1} = \mu \cdot g^t + \frac{{\color{red} \frac{\partial L(f(x_{\text{adv}}^t, \theta), y_{\text{true}})}{\partial x}}}{\left\| {\color{red} \frac{\partial L(f(x_{\text{adv}}^t, \theta), y_{\text{true}})}{\partial x}}\right\|_1}
\end{equation}
\begin{equation}
\small
    x_{\text{adv}}^{t+1} = x_{\text{adv}}^t + \epsilon \cdot \text{sign}(g^{t+1} )
\end{equation}

\textbf{DI-FGSM}: Enhances the transferability of adversarial examples across models by applying random transformations to the inputs before computing the gradient.
\begin{equation}
\small
    x_{\text{adv}}^{t+1} = x_{\text{adv}}^t + \epsilon \cdot \text{sign}\left({\color{red} \frac{\partial L(f(T(x_{\text{adv}}^t), \theta), y_{\text{true}})}{\partial x}} \right)
\end{equation}

\textbf{SINI-FGSM}: Combines scale invariance with the iterative process to improve the effectiveness of the attack in scenarios with varying scales.
\begin{equation}
\small
    g^{t+1} = \mu \cdot g^t + \frac{ {\color{red} \frac{\partial L(f(S(x_{\text{adv}}^t), \theta), y_{\text{true}})}{\partial x}}}{\left\| {\color{red} \frac{\partial L(f(S(x_{\text{adv}}^t), \theta), y_{\text{true}})}{\partial x}}\right\|_1}
\end{equation}
\begin{equation}
\small
    x_{\text{adv}}^{t+1} = x_{\text{adv}}^t + \epsilon \cdot \text{sign}(g^{t+1} )
\end{equation}

In these expressions, $x_{\text{adv}}$ denotes the adversarial example, $x$ is the original input, $L$ is the loss function, $f$ represents the model function, $\theta$ stands for model parameters, $y_{\text{true}}$ is the true label, $\epsilon$ is the perturbation magnitude, $\alpha$ is the step size, $\mu$ is the momentum decay factor, and $T$ and $S$ are the transformation operations used in DI-FGSM and SINI-FGSM, respectively. From the equations corresponding to each method, it is evident that DSP determines the suitable direction for sample updates through the gradient of the loss function with respect to the input, denoted by $\frac{\partial L}{\partial x}$. The remaining parts involve DOP, which utilizes directional information to optimize the samples.

\section{Ablation Results}

\subsection{The Impact of Learning Rate on P3A Performance}

\begin{figure*}[h]
    \centering
    \includegraphics[width=0.73\linewidth]{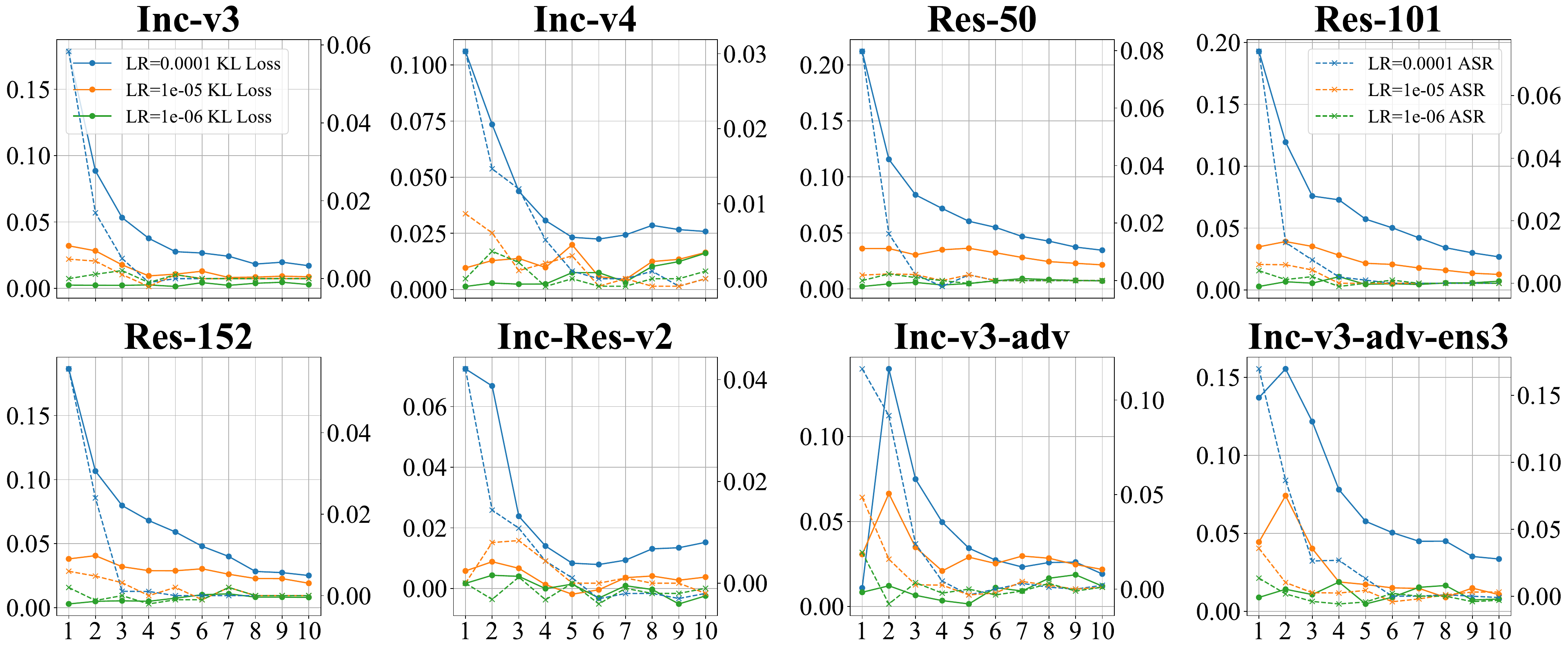}
    \caption{Performance of P3A on various models under different model fine-tuning learning rate. The left y-axis represents the change in KL Loss, and the right y-axis represents the change in ASR}
    \label{fig:LR}
\end{figure*}
In this section, we assessed the impact of different learning rates (LR) for fine-tuning model parameters on the performance of the P3A method. We fixed the attack step size at $\alpha=1$ and experimented with three different learning rates: 0.0001, 0.00001, and 0.000001. The results from Fig.~\ref{fig:LR} indicate that the P3A method achieves the best performance with a learning rate of 0.0001. 


\subsection{The Impact of $\alpha$ on P3A Performance}

\begin{figure*}[t]
    \centering
    \includegraphics[width=0.73\linewidth]{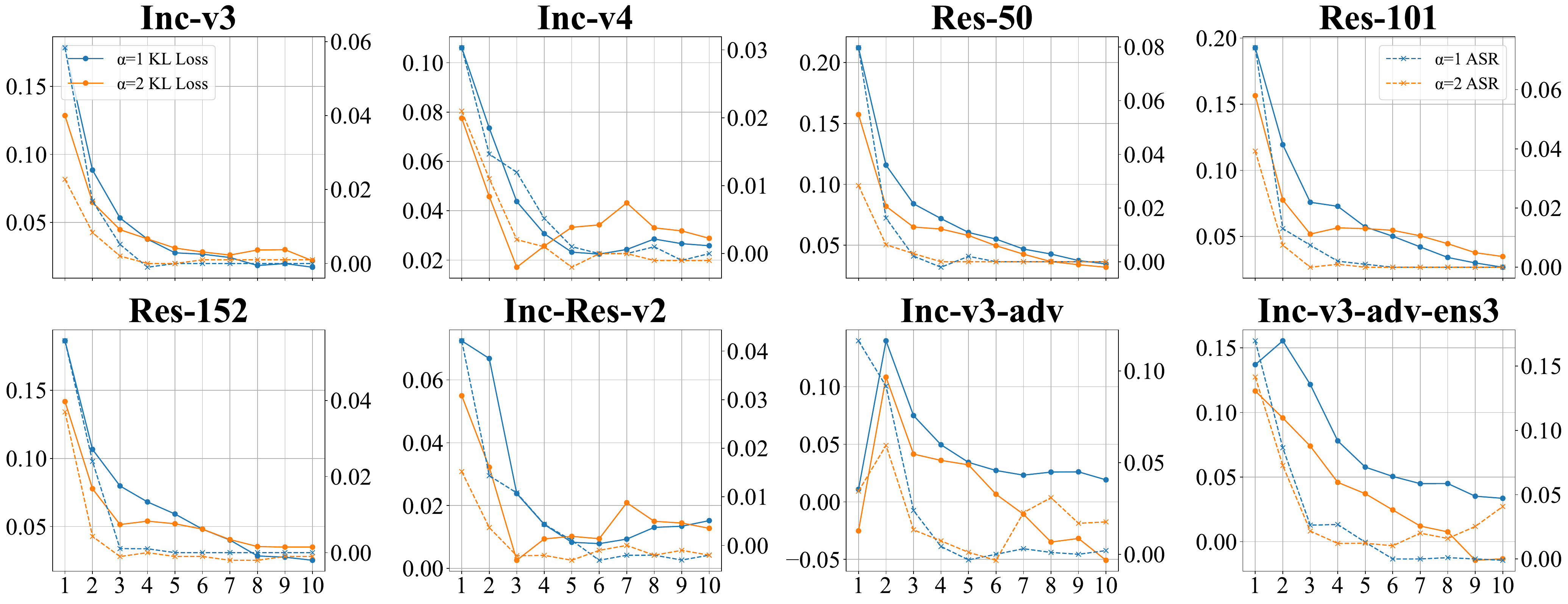}
    \caption{Performance of P3A on various models under different attack step sizes $\alpha$. The left y-axis represents the change in KL Loss, and the right y-axis represents the change in ASR}
    \label{fig:alpha}
\end{figure*}
In this section, we meticulously examine the impact of different attack step sizes $\alpha$ on the performance of the P3A attack method. As illustrated in Fig.~\ref{fig:alpha}, the performance of P3A generally outperforms the case of $\alpha = 2$ when $\alpha = 1$ across most models. $\alpha = 1$ offers a more robust performance and is generally the preferred step size in most scenarios. However, $\alpha = 2$ may exhibit advantages under specific conditions, particularly with an increased number of iterations. Therefore, considering the specificity of the model and the iterative nature of the attack is crucial when designing attack strategies.

\subsection{The Impact of 4 different parameter updates methods on P3A Performance}

\begin{figure*}[t]
    \centering
    \includegraphics[width=\linewidth]{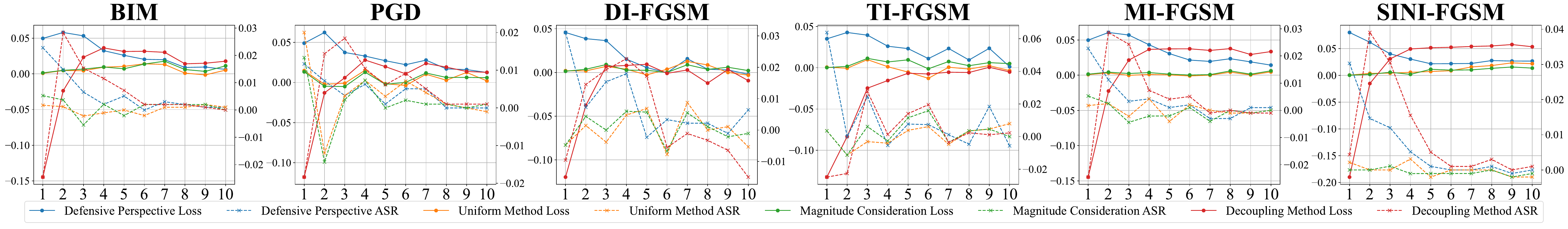}
    \caption{Performance of P3A on various models under 4 different parameter updates methods. The left y-axis represents the change in Cross Entropy Loss, and the right y-axis represents the change in ASR}
    \label{fig:compl}
\end{figure*}
In this section, we investigate the impact of the four different parameter update methods introduced in Sec.~\ref{updates_methods} on the performance of the P3A attack. The attack step size $\alpha$ was fixed at 1, and the learning rate (LR) was set to 0.0001. As shown in Fig.~\ref{fig:compl}, the experimental results indicate that among all the update methods compared, the performance of P3A is optimal when parameters are updated using the Defensive Perspective approach. 

\end{document}